\documentclass[lettersize,journal]{IEEEtran}
\usepackage{amsmath,amsfonts}
\usepackage{multirow}
\usepackage{array}
\usepackage[caption=false,font=normalsize,labelfont=sf,textfont=sf]{subfig}
\usepackage{textcomp}
\usepackage{stfloats}
\usepackage{url}
\usepackage{verbatim}
\usepackage{graphicx}
\usepackage{cite}
\usepackage{subcaption}  
\usepackage{booktabs}    
\usepackage{amsfonts}
\usepackage{pifont}
\newtheorem{definition}{Definition}
\usepackage[ruled,vlined]{algorithm2e}
\begin{document}

\title{GuardFed: A Trustworthy Federated Learning Framework Against Dual-Facet Attacks}

\author{Yanli Li, Yanan Zhou, Zhongliang Guo, Nan Yang, Yuning Zhang, Huaming Chen, Dong Yuan,\\ Weiping Ding*, ~\IEEEmembership{Senior Member,~IEEE}, Witold Pedrycz, ~\IEEEmembership{Life Fellow,~IEEE}

\thanks{Yanli Li is with the School of Artificial Intelligence and Computer Science, Nantong University, Nantong, 226019, China, and also with the School of Electrical and Computer Engineering, The University of Sydney, Sydney, 2006, Australia (e-mail: yanli.li@sydney.edu.au).}
\thanks{Yanan Zhou, Nan Yang, Yuning Zhang, Huaming Chen, and Dong Yuan are the School of Electrical and Computer Engineering, The University of Sydney, Sydney, 2006, Australia (e-mail: yzho5556@uni.sydney.edu.au, \{n.yang, yuning.zhang1, huaming.chen, dong.yuan \}@sydney.edu.au).}
\thanks{Zhongliang Guo is with the School of Computer Science, University of St Andrews, United Kingdom, KY16 9SX (e-mail: zg34@st-andrews.ac.uk).}
\thanks{Weiping Ding  \textit{(Corresponding author)} is the School of Artificial Intelligence and Computer Science, Nantong University, Nantong, 226019, China, and also the Faculty of Data Science, City University of Macau, Macau 999078, China (e-mail: dwp9988@163.com) }
\thanks{Witold Pedrycz is with the Department of Electrical and Computer Engineering, University of Alberta, Edmonton, AB T6R 2V4, Canada (e-mail: wpedrycz@ualberta.ca).}

}

\markboth{Journal of \LaTeX\ Class Files,~Vol.~14, No.~8, August~2021}%
{Shell \MakeLowercase{\textit{et al.}}: A Sample Article Using IEEEtran.cls for IEEE Journals}

\maketitle

\begin{abstract}
Federated learning (FL) enables privacy-preserving collaborative model training but remains vulnerable to adversarial behaviors that compromise model utility or fairness across sensitive groups. While extensive studies have examined attacks targeting either objective, strategies that simultaneously degrade both utility and fairness remain largely unexplored. To bridge this gap, we introduce the Dual-Facet Attack (DFA), a novel threat model that concurrently undermines predictive accuracy and group fairness. Two variants, Synchronous DFA (S-DFA) and Split DFA (Sp-DFA), are further proposed to capture distinct real-world collusion scenarios. Experimental results show that existing robust FL defenses, including hybrid aggregation schemes, fail to resist DFAs effectively. To counter these threats, we propose GuardFed, a self-adaptive defense framework that maintains a fairness-aware reference model using a small amount of clean server data augmented with synthetic samples. In each training round, GuardFed computes a dual-perspective trust score for every client by jointly evaluating its utility deviation and fairness degradation, thereby enabling selective aggregation of trustworthy updates. Extensive experiments on real-world datasets demonstrate that GuardFed consistently preserves both accuracy and fairness under diverse non-IID and adversarial conditions, achieving state-of-the-art performance compared with existing robust FL methods. 
\end{abstract}


\section{Introduction}
The rapid advancement of deep learning (DL) has greatly accelerated the deployment of intelligent automation systems \cite{bangemann2014state}, providing smart services across diverse application domains. Alongside this evolution, there is an increasing emphasis on human-centered values such as privacy, fairness, and security, which extend beyond traditional performance-oriented objectives. Among the candidate technologies that address these concerns, Federated Learning (FL) has emerged as a privacy-preserving and decentralized paradigm \cite{10.1145/3725221}, allowing multiple participants to collaboratively train models without sharing raw data.

Under the FL architecture \cite{luzon2024tutorial}, participants operate with a certain degree of autonomy, each performing local model training and sharing only model gradients during the learning process. A trusted central server coordinates the training by distributing the global model, collecting client updates, and aggregating them to achieve global optimization. However, the distributed nature of FL also introduces a broader attack surface \cite{li2025threats}. Malicious participants may intentionally manipulate their local data or gradient updates to inject poisoned contributions \cite{xie2020fall, li2024contribution, el2022genuinely, damaskinos2019aggregathor}, thereby degrading global model performance or introducing a backdoor to the system. 

Besides model utility, the fairness of the global model has also been identified as an emerging vulnerable target for adversarial manipulation \cite{sheng2024fairguard}. In the context of federated learning, group fairness refers to the requirement that individuals from different sensitive groups, such as those defined by race or gender, should receive similar treatment or prediction outcomes from the model \cite{du2021fairness, rafi2024fairness}. The recent study \cite{sheng2024fairguard} has shown that adversaries can deliberately flip sensitive attributes in their local training data in order to induce significant disparities in model behavior across different groups. These vulnerabilities in both performance and fairness have hindered the full realization of federated learning’s potential for delivering trust worthy and intelligent services in real-world applications.

To defend against malicious behaviors in FL, various robust aggregation algorithms have been proposed, typically relying on statistical techniques \cite{blanchard2017machine, yin2018byzantine} or trusted reference datasets \cite{cao2021fltrust, sheng2024fairguard, li2024contribution} to assess client reliability and exclude adversarial updates. While these methods have shown effectiveness against isolated attacks on either performance or fairness, it is unrealistic to assume that adversaries would limit themselves to a single objective in practice \cite{li2025threats, Kairouz2021AdvancesLearning}. A more strategic and damaging threat involves simultaneously compromising both utility and fairness, yet such dual-objective attacks remain largely unexplored. To address this gap, we introduce the \textbf{Dual-Facet Attack (DFA)}, a novel threat model that jointly degrades predictive accuracy and fairness through a combination of model perturbation and sensitive attribute manipulation. We consider two variants: \textit{Synchronous DFA (S-DFA)}, where all attackers perform both actions, and \textit{Split DFA (Sp-DFA)}, where different groups execute each component independently.

In response, we propose \textbf{GuardFed}, a self-adaptive federated learning system designed to defend against both poisoning and fairness-harming attacks. GuardFed requires the server to collect only a small amount of clean data, known as root data, which can be augmented using generative AI techniques to reach the desired volume. This synthetic dataset allows the server to train a fairness-aware reference model. During training, each client update is evaluated for its trustworthiness from two perspectives: its deviation from the reference model and its group fairness performance evaluated on the root data. A dual-perspective trust score is then computed accordingly, and only updates exceeding a predefined threshold are aggregated into the global model. GuardFed resists compound adversarial behaviors without relying on large-scale server-side data or strong trust assumptions, thereby enabling robust, fair, and privacy-preserving collaborative intelligence.

Extensive experiments across multiple datasets and data distributions are conducted to assess the effectiveness of the proposed Dual-Facet Attack (DFA) and the robustness of the GuardFed system. The results demonstrate that DFA can significantly compromise either model utility or fairness when targeting existing robust FL methods or even hybrid defenses combining performance- and fairness-focused techniques. In contrast, GuardFed effectively mitigates both poisoning and fairness-harming attacks, including DFA. It provides reliable protection for both objectives and maintains strong performance in both adversarial and benign scenarios.

To the best of our knowledge, this is the first work that simultaneously considers both performance and fairness vulnerabilities in federated learning systems. Our contributions are summarized as follows:
\begin{itemize}
    \item We propose \textbf{Dual-Facet Attack} (DFA), a novel threat strategy that simultaneously targets both model utility and fairness. DFA can effectively degrade learning performance and/or fairness, even in the presence of robust FL defenses, posing a significant threat to real-world FL systems.

    \item We design \textbf{GuardFed}, a defense framework that evaluates trust through dual-perspective scoring and performs adaptive aggregation. GuardFed can effectively defend against conventional poisoning attacks and emerging fairness-compromising threats, including DFA.
    
    \item We conduct extensive experiments on widely-used fairness-sensitive datasets under both IID and non-IID settings. The experimental results demonstrate that GuardFed outperforms existing robust FL methods in terms of accuracy, robustness, and fairness across both adversarial and benign environments.
\end{itemize}

The rest of the paper is structured as follows: Section \ref{S2} reviews the existing FL poisoning, fair attacks and defense. Section \ref{S3} setups the federated learning system and group fairness. Section \ref{S4} and \ref{S5} introduce the Dual-Facet Attack (DFA) and GuardFed system. Section \ref{S6} presents the experimental results and analysis. Section \ref{S7} concludes the paper.

\section{Related Work and Research Gap} \label{S2}
In this section, we begin by reviewing the existing vulnerabilities of FL and their corresponding defense frameworks. We then examine recent developments in fairness-aware FL systems, and finally highlight the remaining research gaps identified from prior work.
\subsection{Poisoning Attacks and Byzantine-resilient Aggregations for Federated Learning}
The distributed learning paradigm enables federated learning (FL) to preserve data privacy, allowing participants substantial autonomy to train models locally. However, this loosely coupled structure also exposes FL to a variety of attacks, particularly when the aggregation mechanism is based on linear operations. As demonstrated by Blanchard et al. \cite{blanchard2017machine}, no linear combination of model updates can tolerate even a single Byzantine client, highlighting the inherent vulnerability of linear aggregation in adversarial settings.

Poisoning attacks, as one of the most representative threats in federated learning aimed at degrading global model performance, have been extensively studied in recent years. Straightforward attack strategies include directly reversing model updates (Reverse Attack \cite{damaskinos2019aggregathor}) or replacing a predefined percentage of neural parameters with zeros (Drop Attack \cite{el2022genuinely}) or random values (Random Attack \cite{el2022genuinely}). However, despite their simplicity and effectiveness, these attack strategies are relatively easy to detect and mitigate due to the large deviation they introduce from benign client behaviors \cite{kairouz2021advances}. 

To enhance stealthiness, recent studies \cite{baruch2019little, shejwalkar2021manipulating, xie2020fall} have explored leveraging statistical insights to introduce subtle perturbations, aiming to achieve attack objectives while keeping malicious clients indistinguishable from benign ones. For instance, the Fall Of Empires (FOE) attack \cite{xie2020fall} first computes the coordinate-wise mean of benign gradients and then scales it by a factor slightly greater than 1 to craft the malicious update. Experimental results have demonstrated that the FOE attack can successfully bypass many defense mechanisms and degrade global model performance by up to 20\%. A similar strategy has been explored by FedImp~\cite{li2024fedimp}, which crafts perturbations based on the mean and standard deviation of the attacker’s own neural network parameters.

In response, a variety of Byzantine-resilient aggregation methods have been proposed to replace the simple averaging used in vanilla FL and enhance its robustness. Notable early studies include Krum \cite{blanchard2017machine}, Trimmed Mean \cite{yin2018byzantine}, Median \cite{yin2018byzantine}, and Bulyan \cite{guerraoui2018hidden}, which aim to filter out suspicious client gradients based on statistical measures such as L2 distance, coordinate-wise outliers, etc. While these methods demonstrate effectiveness under IID settings, their defensive performance significantly degrades in non-IID scenarios, particularly when attack strategies become more adaptive and stealthy \cite{li2025threats}. This is because statistical methods rely on gradient similarity to detect outliers, an assumption that holds in IID settings but breaks down under non-IID distributions. In such cases, natural gradient divergence masks malicious behavior, especially when attackers employ small, stealthy perturbations.

Recently, a new research paradigm centered on evaluation-based robust aggregation has emerged \cite{cao2021fltrust, park2021sageflow, li2024contribution, kabir2024flshield}, demonstrating enhanced resilience compared to traditional statistical methods. These methods typically rely on a clean evaluation dataset, which is usually curated by the server, to assess client gradients from multiple perspectives. For instance, FLTrust \cite{cao2021fltrust} requires the server to maintain a server-side model trained on trusted data, and measures the cosine similarity between this reference model and each client update to determine whether the client is benign. Similarly, FLEvaluate \cite{guo2023flevaluate} and Class-Balanced FL \cite{li2024contribution} utilize server-collected data to directly evaluate the learning performance of client models, with the shared goal of improving robustness against malicious manipulation.

\subsection{Fairness-aware Federated Learning}
Fairness is a critical requirement in human-centric smart services, where trained models are expected to deliver consistent performance across different demographic groups, particularly with respect to sensitive attributes such as gender and race \cite{huang2024federated, rafi2024fairness, he2025towards}. To address this challenge, studies \cite{du2021fairness, ray2022fairness}  formulate fairness in federated learning as a constrained optimization problem, in which fairness requirements are incorporated as explicit constraints during the model training process. FairFed \cite{ezzeldin2023fairfed} introduces a reweighting technique, adapted from traditional machine learning, into the federated learning setting. The method restructures the loss function to promote fairness in the global model, particularly under training datasets that exhibit “equal sampling” across sensitive groups. FairEM \cite{zhang2025sffl} aims to ensure that clients obtain fair models while preserving the privacy of their sensitive feature information. Specifically, FairEM \cite{zhang2025sffl} decomposes each client’s training objective into fairness-aware sub-objectives based on the underlying data distribution, thereby improving both local fairness and overall model performance.

While most prior studies \cite{ray2022fairness, ezzeldin2023fairfed, zhang2025sffl} on fairness-aware federated learning assume a benign training environment, a recent work \cite{sheng2024fairguard} is among the first to address this problem under adversarial conditions, where attackers can deliberately manipulate sensitive attributes, for example, flipping a client’s gender label from “female" to “male." To defend against such fairness-harming attacks, FairGuard \cite{sheng2024fairguard} has been proposed. By leveraging clustering results obtained from evaluations on a randomly generated dataset, FairGuard is able to restore model fairness to a level comparable to that of an unattacked system.\\

\textbf{Research Gap:} Despite extensive studies on attacks and robust aggregation in federated learning, existing works predominantly focus on either degrading model utility or manipulating fairness, while overlooking the possibility of adversaries targeting both simultaneously. To the best of our knowledge, no prior work has investigated such dual-objective attacks, nor proposed effective defense mechanisms against them. In practice, however, it is unrealistic to assume that attackers would limit themselves to a single objective, especially when jointly harming model performance and fairness can yield greater systemic disruption. Furthermore, as we will show in our experiments, directly combining fairness-aware FL frameworks with robust aggregation techniques fails to provide sufficient protection under coordinated dual-facet attacks. These observations motivate the development of a new attack strategy, Dual-Facet Attack (DFA), and its dedicated defense framework GuardFed, both of which will be detailed in Sections~\ref{S4} and~\ref{S5}. 

We summarize this research gap in Table~\ref{comparison}, which contrasts the threat coverage of our proposed Dual-Facet Attack and GuardFed system with representative prior works.

\begin{table}[htbp]\footnotesize
  \centering
  \caption{Comparison of the applicability scope of Dual-Facet Attack and GuardFed System (Per:performance, Fair:fairness).}
  \label{comparison}
    \centering
    \begin{tabular}{p{2cm} p{0.5cm} p{0.5cm}| p{2.2cm} p{0.5cm} p{0.5cm}}
      \hline\hline
      \textbf{Attack Methods} & \textbf{Perf} & \textbf{Fair} &\textbf{Defense Methods} & \textbf{Perf} & \textbf{Fair} \\
      \hline
      \cite{shejwalkar2021manipulating}\hfill \textit{NDSS’21} & \ding{51} & \ding{55}&\cite{park2021sageflow}\hfill \textit{NeurIPS’21} & \ding{51} & \ding{55} \\
      \cite{el2022genuinely}\hfill \textit{DisCom’22} & \ding{51} & \ding{55}&\cite{cao2021fltrust}\hfill \textit{NDSS’21} & \ding{51} & \ding{55} \\
      \cite{wei2023covert}\hfill \textit{TDSC’23} & \ding{51} & \ding{55} & \cite{guo2023flevaluate}\hfill \textit{TruCom’23} & \ding{51} & \ding{55}\\
      \cite{li2024fedimp}\hfill \textit{COSE’24} & \ding{51} & \ding{55} &\cite{li2024contribution}\hfill \textit{InfoSci’24} & \ding{51} & \ding{55} \\
      \cite{meerza2024eab}\hfill \textit{IJCAI’24} & \ding{55} & \ding{51} & \cite{yazdinejad2024robust}\hfill \textit{TIFS’24} & \ding{51} & \ding{55}\\
      \cite{sheng2024fairguard}\hfill \textit{TDSC’25} & \ding{55} & \ding{51} & \cite{sheng2024fairguard}\hfill \textit{TDSC’25} & \ding{55} & \ding{51}\\
      \hline
      \textbf{Dual-Facet} & \ding{51} & \ding{51} & \textbf{GuardFed} & \ding{51} & \ding{51}\\
      \hline
    \end{tabular}

\end{table}

\section{Problem Setup} \label{S3}
\subsection{Federated Learning}

We consider a standard cross-device Horizontal Federated Learning (HFL) setting (clients share the same feature space but have different samples) with a central server coordinating $N$ clients. The server maintains a global model $M_G^t \in \mathbb{R}^d$ at communication round $t$, while each client $n \in \{1, \dots, N\}$ possesses a private dataset $\mathcal{D}_n$ drawn from a non-independent and identical distribution (non-IID).

At the beginning of each round $t$, the server broadcasts the global model $M_G^t$ to all selected clients. Each client initializes its local model with $M_G^t$ and performs local training for $E$ epochs using stochastic gradient descent or a similar optimizer. The local update process seeks to minimize the empirical risk over the client's private dataset:
\begin{equation}
    M_n^t = \arg\min_{M} \ \mathcal{L}_n(M) = \frac{1}{|\mathcal{D}_n|} \sum_{(x,y) \in \mathcal{D}_n} \mathcal{L}(M(x), y),
\end{equation}
where $\mathcal{L}(\cdot)$ is the loss function (e.g., cross-entropy). After local training, the model update is computed as:
\begin{equation}
    g_n^t = M_n^t - M_G^t.
\end{equation}

Upon receiving updates from the selected clients, the server performs aggregation to obtain the new global model. In standard FedAvg, the server performs a weighted average over all client updates, where $\eta$ is the global learning rate.:
\begin{equation}
    M_G^{t+1} = M_G^t + \eta \cdot \sum_{n=1}^N \frac{|\mathcal{D}_n|}{\sum_{j=1}^N |\mathcal{D}_j|} \cdot g_n^t,
\end{equation}

This decentralized training protocol allows collaborative learning without centralized access to raw data. However, this paradigm is vulnerable to unreliable or adversarial updates. Clients may submit poisoned or unfair models, either unintentionally due to local bias or deliberately to mislead the global model. Therefore, additional mechanisms are required to evaluate and filter client contributions based on their utility and fairness, which is the motivation behind the GuardFed framework.

\subsection{Group Fairness}
To quantify group-level disparities, we adopt two absolute fairness metrics: \textit{Absolute Statistical Parity Difference (ASPD)} and \textit{Absolute Equal Opportunity Difference (AEOD)}, which are widely used in FL context. These metrics capture the extent to which a model exhibits differing behaviors across sensitive groups.

\begin{definition}[\textbf{Absolute Statistical Parity Difference (ASPD) \cite{zemel2013learning}}]
Let $a \in \{0, 1\}$ be a binary sensitive attribute (e.g., gender), and let $M(x)$ denote the model prediction. The Absolute Statistical Parity Difference is defined as:
\begin{equation}
\begin{split}
\text{ASPD}(M) = \big| 
&\Pr(M(x)=1 \mid a=0) \\
&- \Pr(M(x)=1 \mid a=1)
\big|.
\end{split}
\end{equation}
\end{definition}
ASPD measures the absolute deviation in the likelihood of receiving a positive prediction between sensitive groups, irrespective of the true label. A smaller ASPD implies that the model makes group-independent predictions.

\begin{definition}[\textbf{Absolute Equal Opportunity Difference (AEOD) \cite{zemel2013learning}}]
Let $y \in \{0, 1\}$ be the true label. The Absolute Equal Opportunity Difference is defined as:
\begin{equation}
\begin{split}
\text{AEOD}(M) = \big| 
&\Pr(M(x)=1 \mid a=0, y=1) \\
&- \Pr(M(x)=1 \mid a=1, y=1) 
\big|.
\end{split}
\end{equation}
\end{definition}
AEOD captures the absolute difference in true positive rates between sensitive groups. A lower AEOD value indicates that qualified individuals across groups are treated more equally by the model.

\section{Dual-Facet Attack} \label{S4}

In this section, we propose \textit{Dual-Facet Attack (DFA)} to simultaneously target both the performance and fairness of federated learning systems. DFA comprises two adversarial operations: (1) flipping the sensitive attribute $a$ in the client’s local dataset to disrupt group fairness, and (2) perturbing model updates with adversarial noise to degrade global accuracy.

We consider a federated learning setting where a fixed proportion $\alpha$ of participating clients are malicious. These adversaries have complete control over their local training procedures and uploaded model updates, including flipping sensitive attributes or injecting noise into gradients. They are allowed to communicate and coordinate with each other to launch synchronized or role-divided attacks, including our proposed Dual-Facet Attack.   

Let $g_n^t$ be the local update from client $n$ at round $t$, and let $\mathcal{A}_{\text{fair}}(\cdot)$ and $\mathcal{A}_{\text{perf}}(\cdot)$ represent the fairness attack and performance attack operators, respectively. Here, we use Gaussian noise injection ($\delta \sim \mathcal{N}(0, \sigma^2 I)$ \cite{hossain2025exploiting}) to instantiate $\mathcal{A}_{\text{perf}}$, though other poisoning strategies may also be applied. By combining these in different ways, DFA results in two variant attack strategies, which are described as follows.
 
\begin{definition}[\textbf{Synchronous Dual-Facet Attack (S-DFA)}]
In this variant, each malicious client simultaneously applies both the fairness and performance attack. That is, all adversarial clients manipulate their local dataset by flipping the sensitive attribute $a \leftarrow 1 - a$, retrain their local model on the perturbed data, and then inject noise into the resulting model update. Formally, the poisoned update is:
\begin{equation}
    \widetilde{g}_n^t = \mathcal{A}_{\text{perf}} \left( \mathcal{A}_{\text{fair}}(g_n^t) \right) = g_n^t + \delta, \quad \delta \sim \mathcal{N}(0, \sigma^2 I),
\end{equation}
\end{definition}

\begin{definition}[\textbf{Split Dual-Facet Attack (Sp-DFA)}]
In this variant, the malicious clients are divided into two disjoint subgroups. One subgroup performs fairness attacks ($\mathcal{A}_{\text{fair}}$) by flipping the sensitive attribute in their local dataset and training on the altered data. The other subgroup keeps the original data unchanged but injects Gaussian noise into the model updates ($\mathcal{A}_{\text{perf}}$). Formally, the poisoned updates are:
\begin{equation}
\widetilde{g}_n^t =
\begin{cases}
\mathcal{A}_{\text{fair}}(g_n^t), & \text{if } n \in \mathcal{A}_{\text{fair}},\\
\mathcal{A}_{\text{perf}}(g_n^t) = g_n^t + \delta, \quad \delta \sim \mathcal{N}(0, \sigma^2 I), & \text{if } n \in \mathcal{A}_{\text{perf}}.
\end{cases}
\end{equation}
Sp-DFA represents a coordination strategy in which heterogeneous adversaries collaboratively compromise both utility and fairness objectives.
\end{definition}

\section{GuardFed System Design} \label{S5}
\begin{figure*} 
    \centering
    \includegraphics[width=1.0\linewidth]{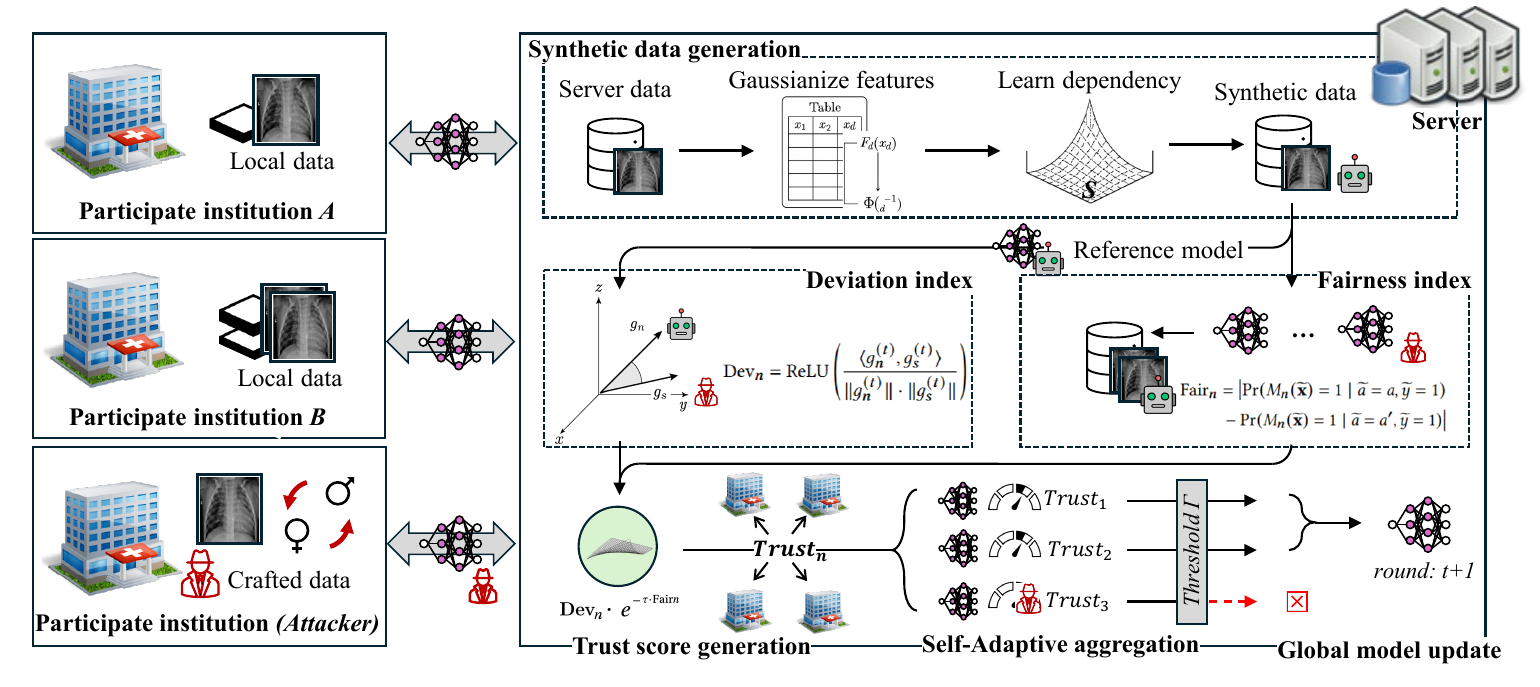}
    \caption{Overview of the GuardFed system. All participants, including potential attackers, perform local model training and upload their updates. The server applies a self-adaptive aggregation mechanism to ensure both fairness and robustness in the global model.}
    \label{overview}
\end{figure*}

In this section, we introduce the design of the GuardFed system, which aims to defend both fairness and performance in adversarial federated learning environments. As phrased by study \cite{blanchard2017machine}, traditional linear aggregation schemes are highly vulnerable, even a single Byzantine client can significantly degrade global performance. To address this issue, GuardFed dynamically evaluates the trustworthiness of each client to determine whether their update should be included in the aggregation process. A synthetic dataset maintained on the server is utilized to support the generation of fairness and deviation indices, which are jointly used to compute an honesty score for each client and guides the self-adaptive aggregation process. Fig. \ref{overview} illustrates an overview of the proposed GuardFed system.

The next subsection begins with a detailed explanation of each core component of GuardFed, followed by a description of the complete learning pipeline.

\subsection{Synthetic Data Generation}

To evaluate the honesty of client models, a mainstream approach is to maintain a self-collected clean dataset on the server for performance assessment. While numerous studies \cite{cao2021fltrust,park2021sageflow, li2024contribution} have confirmed the importance of a sufficiently large validation set for detecting malicious clients, relying on a single server to collect such data is often infeasible in practical federated learning deployments. To fully exploit the potential of the evaluation dataset in real-world scenarios, GuardFed leverages generative AI techniques to synthesize data on the server side based on a small number of existing samples. Considering that the most fairness-sensitive federated learning tasks utilize tabular data, we introduce a Gaussian Copula-based method \cite{patki2016synthetic, sklar1959fonctions} for synthetic tabular data generation. This approach, however, can be extended to other generative techniques (e.g., LLM or GAN) when the training data is non-tabular. As this study does not focus on generating high-quality synthetic data, other methods in this area are not discussed in detail here.

Given a standalone table with $r$ rows and $d$ columns, let $\textbf{x}^{(i)} = (x^{(i)}_1, x^{(i)}_2, ..., x^{(i)}_d) \in \mathbb{R}^d$ denote the $i$-th sample in the original dataset $\mathcal{D} = \{\textbf{x}^{(1)}, ..., \textbf{x}^{(r)}\}$. The Gaussian Copula model generates synthetic samples by decoupling the modeling of marginal distributions from inter-feature dependencies. Specifically, the generation process consists of the following steps:

We first model the marginal and joint properties of the tabular data separately. For each feature $x^{(i)}_d$, we begin by transforming it into a uniform variable $u^{(i)}_d \in [0,1]$ using its empirical marginal cumulative distribution function $F_d$:
\begin{equation}
    u^{(i)}_d = F_d(x^{(i)}_d).
\end{equation}

To remove distributional biases and normalize the variables, we then apply the inverse standard Gaussian CDF $\Phi^{-1}$ (i.e., the probit function) to obtain a Gaussianized variable $z^{(i)}_d$:
\begin{equation}
    z^{(i)}_d = \Phi^{-1}(u^{(i)}_d) = \Phi^{-1}(F_d(x^{(i)}_d)),
\end{equation}
which transforms the original data sample into $\textbf{z}^{(i)} = (z^{(i)}_1, ..., z^{(i)}_d)$. After mapping all $n$ samples to the Gaussian space, we compute the empirical covariance matrix $\Sigma \in \mathbb{R}^{d \times d}$ over $\{\textbf{z}^{(i)}\}_{i=1}^n$, which captures the inter-feature dependencies in a scale-invariant manner.

Next, synthetic samples are drawn from the learned multivariate Gaussian distribution: 
\[
\widetilde{\textbf{z}} = (\widetilde{z}_1, ..., \widetilde{z}_d) \sim \mathcal{N}(0, \Sigma).
\]
Each sampled component $\widetilde{z}_d$ is mapped back to the uniform space via the standard Gaussian CDF:
\begin{equation}
    \widetilde{u}_d = \Phi(\widetilde{z}_d),
\end{equation}
and subsequently transformed into the original data space using the inverse marginal distribution $F_d^{-1}$:
\begin{equation}
    \widetilde{x}_d = F_d^{-1}(\widetilde{u}_d), \quad \widetilde{\textbf{x}} = (\widetilde{x}_1, ..., \widetilde{x}_d).
\end{equation}

The resulting vector $\widetilde{\textbf{x}}$ represents one synthetic sample that preserves the marginal distributions and correlation structure of the original dataset. Repeating this procedure yields a synthetic dataset $\widetilde{\mathcal{D}}$ that can be reused throughout the federated learning process.

\subsection{Trust Score Generation}

To detect whether clients have manipulated their model updates that undermines the performance or fairness of the global model, GuardFed computes a trust score for each received update using the synthetic dataset $\widetilde{\mathcal{D}}$. This trust score integrates two components: a \textit{fairness index} and a \textit{deviation index}, which respectively assess the update's potential harm from fairness and utility perspectives.

\textbf{Fairness Index}: Upon receiving a client model $M_n$, the server evaluates its fairness using the synthetic dataset $\widetilde{\mathcal{D}} = \{(\widetilde{\textbf{x}}^{(i)}, \widetilde{a}^{(i)}, \widetilde{y}^{(i)})\}_{i=1}^{I}$, where $\widetilde{a}$ denotes the sensitive attribute and $\widetilde{y}$ is the target label. Formally, the Fairness Index for client $n$ is computed as:
\begin{equation}
\label{eq:fairness_index}
\begin{split}
\text{Fair}_n = \big|
&\Pr(M_n(\widetilde{\mathbf{x}})=1 \mid \widetilde{a}=a, \widetilde{y}=1) \\
&- \Pr(M_n(\widetilde{\mathbf{x}})=1 \mid \widetilde{a}=a', \widetilde{y}=1)
\big|, \quad \text{Fair}_n \in [0,1].
\end{split}
\end{equation}
Here, $(a, a')$ denotes two distinct values of the sensitive attribute (e.g., male/female). A higher $\text{Fair}_n$ indicates greater unfairness, i.e., a stronger disparity between subgroups with the same ground-truth label.

\textbf{Deviation Index}:
To guide the global model in a fair and correct learning direction, the server maintains a reference model $M_s$ trained on the synthetic dataset $\widetilde{\mathcal{D}}$ using a reweighting strategy \cite{kamiran2012data}. The weight $w_i$ assigned to each training sample $(\widetilde{\textbf{x}}^{(i)}, \widetilde{a}^{(i)}, \widetilde{y}^{(i)})$ is computed based on its joint and marginal group-label probabilities:
\begin{equation}
    w_i = \frac{\Pr(\widetilde{a}^{(i)}) \cdot \Pr(\widetilde{y}^{(i)})}{\Pr(\widetilde{a}^{(i)}, \widetilde{y}^{(i)})}
\end{equation}
This reweighting ensures that the sensitive attribute $\widetilde{a}$ becomes statistically independent of the label $\widetilde{y}$ under the adjusted distribution, i.e., $\widetilde{a} \perp \widetilde{y}$, thereby mitigating group-level discrimination. Thus, the reference model is trained by minimizing the weighted empirical loss:
\begin{equation}
    M_s^{(t)} = \arg\min_{M} \sum_{i=1}^{I} w_i \cdot \mathcal{L}(M(\widetilde{\textbf{x}}^{(i)}), \widetilde{y}^{(i)})
\end{equation}

During the $t$-th communication round, after collecting client model $M_n^{(t)}$, the server compares its gradient update with that of the reference model $M_s^{(t)}$. Let $g_n^{(t)} = M_n^{(t)} - M_G^{(t-1)}$ and $g_s^{(t)} = M_s^{(t)} - M_G^{(t-1)}$ denote the gradient vectors. The deviation index is calculated as the ReLU-activated cosine similarity:
\begin{equation}
\label{eq:cosine_similarity}
    \text{Dev}_n = \text{ReLU}\left( \frac{\langle g_n^{(t)}, g_s^{(t)} \rangle}{\|g_n^{(t)}\| \cdot \|g_s^{(t)}\|} \right)
\end{equation}
This measures the alignment of the client's update with the fair direction indicated by the reference model. A low (or zero, after ReLU clipping) similarity indicates potential deviation or poisoning behavior.

\textbf{Trust Score:} The final trust score $\text{Trust}_n$ for client $n$ is computed by combining the deviation index $\text{Dev}_n$ and the fairness index $\text{Fair}_n$ as follows:
\begin{equation}
\text{Trust}_n = \text{Dev}_n \cdot e^{- \tau \cdot \text{Fair}_n} \label{16}
\end{equation}
where $\tau \geq 0$ is a tunable hyperparameter that controls the impact of fairness violations on the final score. The fairness index $\text{Fair}_n \in [0,1]$ quantifies the degree of unfairness in the client's model; a larger value indicates more severe group-level disparity. Since our goal is to penalize unfair updates, we apply a negative exponent to the fairness term so that higher unfairness leads to lower trust.

Here, we adopt the exponential formulation to guarantee client trust should remain stable under small fairness fluctuations but decrease sharply when severe violations occur. 
The exponential decay term $e^{-\tau \cdot \text{Fair}_n}$ provides a smooth and bounded scaling factor within $(0,1]$, ensuring numerical stability during iterative aggregation. 
Mild fairness variations are tolerated since maintaining model usability and convergence takes precedence, whereas larger violations are strictly penalized through the tunable coefficient $\tau$.

\subsection{Self-Adaptive Aggregation}

Based on the computed trust scores, the server performs self-adaptive aggregation by selectively including clients whose updates are deemed trustworthy. Specifically, in each communication round $t$, only clients with $\text{Trust}_n > \Gamma$ are aggregated into the global model. The aggregation is performed according to the following rule:
\begin{equation}
    M_G^{t+1} = M_G^{t} + \eta \cdot \frac{1}{|\mathcal{S}^t|} \sum_{n \in \mathcal{S}^t} g_n^t,
\end{equation}
where $M_G^{t}$ denotes the global model at round $t$, $\eta$ is the learning rate, $g_n^t$ is the model update from client $n$, and $\mathcal{S}^t = \{n \mid \text{Trust}_n > \Gamma\}$ is the set of selected clients. This strategy ensures that only those clients contributing both fair and reliable updates influence the global model learning.

\subsection{Full Learning Step of GuardFed}

In each round of the GuardFed system, the server first broadcasts the current global model to all participating clients. Upon receiving the model, each client performs local training on its private dataset and uploads the model update to the server. The server then evaluates each client's update using a synthetic validation set constructed via Gaussian Copula modeling. Specifically, it computes a fairness index that captures the group-level disparity in model predictions, and a deviation index that quantifies the alignment between the client's update and a reference update trained under fairness-aware reweighting.

These two indexes are combined into a unified trust score, and only clients whose trust scores exceed a predefined threshold are selected for aggregation. The global model is then updated by averaging the updates from these selected clients. This mechanism ensures that only updates aligned with both performance and fairness objectives contribute to the global model. The process is repeated iteratively over multiple communication rounds until convergence.

The full learning steps are summarized in Algorithm \ref{algorithm}.

\begin{algorithm}[t]
\caption{GuardFed Learning Step}\label{algorithm}
\KwIn{Initial global model $M_G^0$, total rounds $T$, trust threshold $\Gamma$, balance factor $\tau$, learning rate $\eta$}
\KwOut{Final global model $M_G^T$}

Generate synthetic dataset $\widetilde{\mathcal{D}}$ via Gaussian Copula modeling\;
Train server-side reference model $M_s$ on $\widetilde{\mathcal{D}}$ with fairness-aware reweighting\;

\For{$t = 1$ to $T$}{
    Broadcast global model $M_G^t$ to all clients\;

    \ForEach{client $n \in \{1, ..., N\}$}{
        Local training to obtain updated model $M_n^t$\;
        Compute model update $g_n^t = M_n^t - M_G^t$\;
        Send $M_n^t$ or $g_n^t$ to server\;
    }

    \ForEach{received client model $M_n^t$}{
        Compute fairness index $\text{Fair}_n$ on $\widetilde{\mathcal{D}}$\;
        Compute deviation index $\text{Dev}_n$ by comparing $g_n^t$ with $g_s^t$\;
        Compute trust score: $\text{Trust}_n = \text{Dev}_n \cdot e^{-\tau \cdot \text{Fair}_n}$\;
    }

    Select trusted clients: $\mathcal{S}^t = \{n \mid \text{Trust}_n > \Gamma\}$\;

    Update global model:

    $M_G^{t+1} = M_G^t + \eta \cdot \frac{1}{|\mathcal{S}^t|} \sum_{n \in \mathcal{S}^t}g_n^t$
    
}

\Return{$M_G^T$}
\end{algorithm}

\section{Experiments} \label{S6}
This section presents a series of experiments to evaluate the effectiveness of the proposed Dual-Facet Attack and the GuardFed defense framework. The main experimental objectives are threefold: (i) to demonstrate that the Dual-Facet Attack can significantly degrade either model utility or fairness (or both) when applied to representative robust federated learning systems; (ii) to show that the GuardFed system can effectively defend against various threats, including Dual-Facet Attacks, and maintain both learning performance and fairness; and (iii) to validate that GuardFed also maintains competitive performance and fairness in non-adversarial settings.

In addition, we conduct further investigations into the impact of key hyperparameters and deployment settings to gain deeper insights into the robustness and generalizability of the GuardFed framework.

\subsection{Experimental Settings}
We conduct our experiments on the COMPAS \cite{dressel2018accuracy} and Adult \cite{ding2021retiring} datasets, which are commonly adopted in fairness-aware federated learning studies \cite{ezzeldin2023fairfed, sheng2024fairguard, zeng2023federated, rafi2024fairness}. These datasets are associated with real-world applications such as recidivism prediction and income classification. Both datasets consist of thousands of samples, each described by multiple attributes, among which certain features are designated as sensitive attributes. These sensitive features divide individuals into demographic groups, such as race (e.g., African-American vs. Others) in COMPAS and gender (e.g., female vs. male) in Adult. A multi-layer perceptron (MLP) model is used for training, with a learning rate $\eta=0.005$, learning round $ \in [50, 100]$ and batch sizes $\in[16, 256]$.

We consider a federated learning setting where 20 clients are randomly selected from a pool of 100 candidates in each communication round. Among them, 20\% are controlled by adversaries to perform attacks. The impact of varying the proportion of malicious clients will be further explored in Subsection~6.3. To control the degree of data heterogeneity across clients, we adopt the Dirichlet distribution, where $\alpha = 5000$ corresponds to an IID scenario and $\alpha = 5$ represents a highly non-IID distribution.

To remain stealthy, Dual-Facet attackers adopt the FOE attack strategy ($\mathcal{A}_{\text{perf}}$) to degrade model performance, and manipulate (i.e., flip) sensitive attributes ($\mathcal{A}_{\text{fair}}$) to impair fairness. Two variants, S-DFA and Sp-DFA, are simulated to evaluate the effectiveness of different coordination strategies. Following the GuardFed design, the server acts as the defender, holding 1\% of the original training data and generating an additional 4\% of synthetic samples, resulting in a total of 5\% root data used for defense purposes.

We use FedAvg and FairFed as baselines, and include Median, FLTrust, Class-Balance FL, and FairGuard as representative comparison methods. As introduced in Section~2, Median, FLTrust, and Class-Balance FL are the representative and state-of-the-art Byzantine-resilient FL algorithms targeting performance robustness, while FairGuard focuses on fairness preservation under adversarial conditions. To ensure comparability across methods, we introduce Reweighting technology \cite{kamiran2012data} for all the FL algorithms to ensure their fairness has been enhanced. We further consider a hybrid defense that directly combines FLTrust and FairGuard, in order to examine whether a simple integration of fairness-preserving and robust aggregation strategies can simultaneously address both performance and fairness threats. Specifically, the hybrid defense first employs FairGuard to select candidate clients for aggregation, followed by FLTrust to assign aggregation weights. 

\subsection{Experimental Results}
In this subsection, we present experimental results to validate that the proposed Dual-Facet Attack strategy and the GuardFed system successfully achieve the three designated objectives. We use Accuracy (ACC), Absolute Equalized Odds Difference (AEOD), and Absolute Statistical Parity Difference (ASPD) as evaluation metrics to assess the global model's learning performance and fairness across different groups. Higher accuracy and lower AEOD or ASPD values indicate better overall robustness and fairness of the model.

It is important to emphasize that \textbf{AEOD and ASPD cannot independently reflect fairness}. These metrics should be interpreted alongside the model's accuracy because models with low accuracy may appear deceptively fair. For instance, when a model completely fails and yields 50\% accuracy for both groups in a binary classification task, the fairness metric may misleadingly report a perfect score, such as AEOD of zero. However, this model is entirely non-functional. To ensure a fair and meaningful evaluation, we set accuracy thresholds of 60\% for the COMPAS learning task and 80\% for the ADULT learning task. The fairness of a model is considered valid only when its accuracy exceeds these thresholds.

\begin{figure*} [t]
    \centering
    \includegraphics[width=1.0\linewidth]{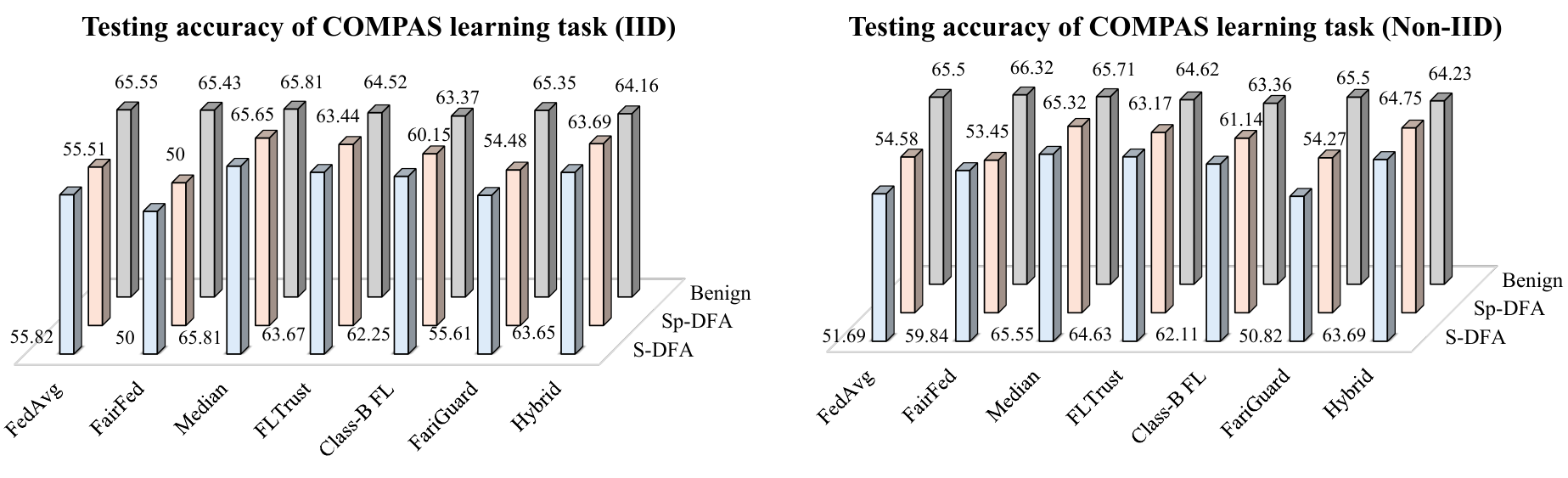}
\caption{Learning performance of fairness-enhanced, robust, and hybrid FL algorithms under benign and Dual-Facet Attack (DFA) scenarios in IID and non-IID settings on the COMPAS dataset.}
    \label{acc}
\end{figure*}

\textbf{(i) DFA attacks can significantly degrade either the learning performance or fairness of existing robust FL algorithms, including hybrid methods.} In terms of utility, DFA reduces the testing accuracy of fairness-enhanced approaches such as FedAvg, FairFed, and FairGuard from approximately 65\% to 55\% on the COMPAS dataset under both IID and non-IID settings. Notably, FairFed experiences the most severe degradation, with accuracy dropping to 50\% in both scenarios, indicating a complete break. Additionally, both DFA variants (Sp-DFA and S-DFA) show comparable effectiveness in degrading model accuracy, with accuracy differences across most FL algorithms being less than 2\%. Figure \ref{acc} illustrates the testing accuracy of existing FL methods under various settings on the COMPAS dataset, while full experimental results and on other datasets will be shown later in Tables \ref{tab:3} and \ref{tab:4}.

\begin{figure*} []
    \begin{flushleft} 
    \includegraphics[width=1.0\linewidth]{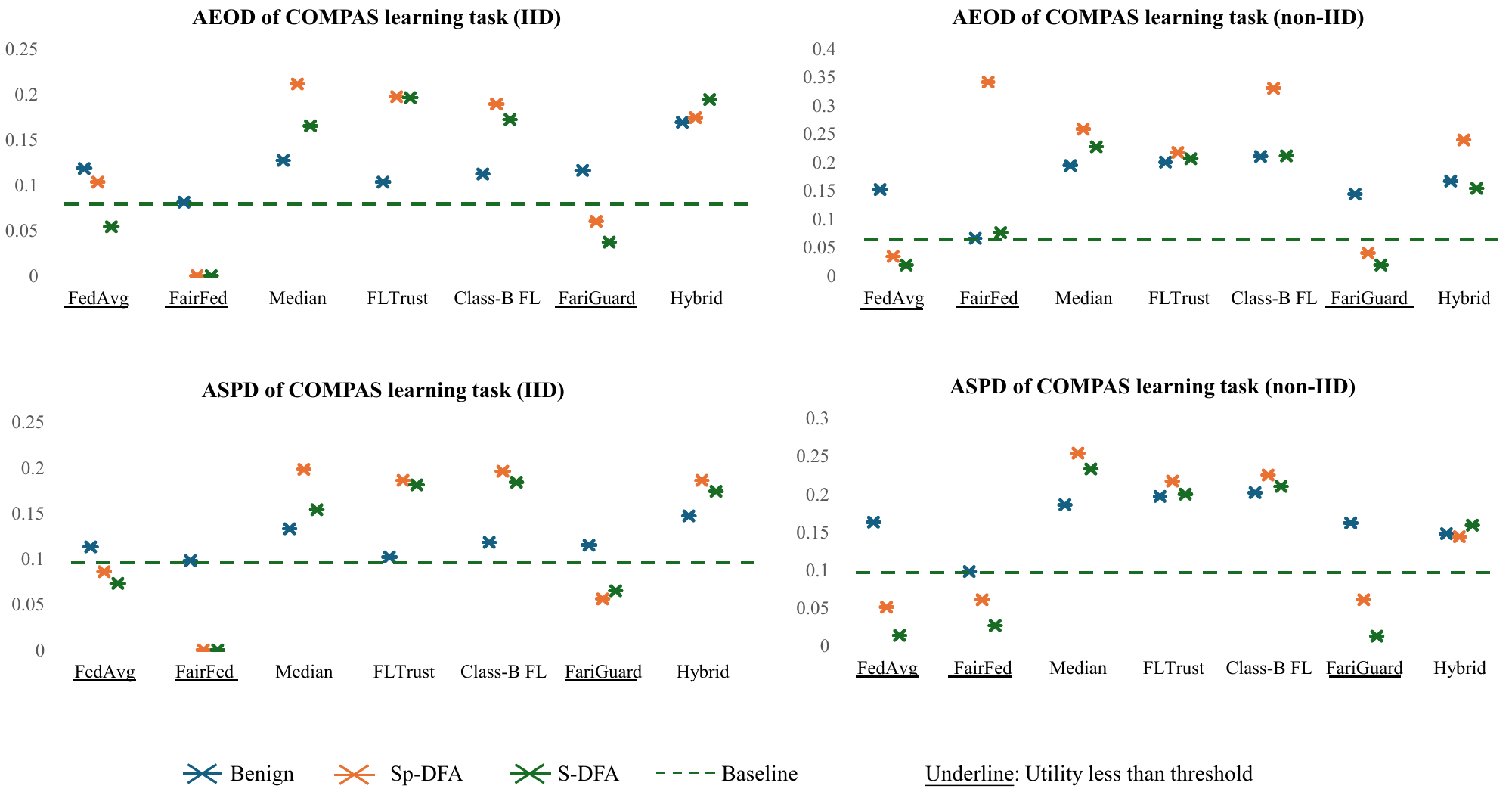}
    \caption{Fairness index (AEOD and ASPD) of fairness-enhanced, robust, and hybrid FL algorithms under benign and Dual-Facet Attack (DFA) scenarios on the COMPAS dataset, evaluated in both IID and non-IID settings. Underlines indicate models whose test accuracy falls below the predefined threshold, rendering the fairness metrics unreliable. }
    \label{eod}
    \end{flushleft} 
\end{figure*}

On the other hand, DFA also proves effective in compromising fairness in robust FL algorithms such as Median, FLTrust, and Hybrid. In particular, Sp-DFA exhibits a stronger fairness-poisoning effect than S-DFA. For example, in the IID setting, Sp-DFA increases the AEOD and ASPD of Median from 0.127 to 0.211 and 0.160, respectively; in the non-IID setting, the metrics rise from 0.194 to 0.258 and 0.227. Notably, although the Hybrid algorithm integrates both fairness- and performance-oriented defenses and is expected to achieve robustness on both fronts, it still suffers from increased AEOD and ASPD under DFA. Furthermore, in various settings, it shows significant unfairness (bias around 0.15–0.2 across groups) compared to the baseline FairFed in benign scenarios. Figure \ref{eod} illustrates the fairness indices of existing FL methods under various settings on the COMPAS dataset, while full experimental results and on other datasets will be shown later in Tables \ref{tab:3} and \ref{tab:4}.
\begin{table*}\small
    
    \centering

\caption{Comparison of the learning performance of existing robust FL methods and GuardFed on the COMPAS dataset. Bold values indicate the best results, while underlines denote fairness scores when the accuracy exceeds the predefined threshold.}
    \makebox[\textwidth][l]{%
    \begin{tabular}{c|c|l|ccccc|ccccc}    
    \hline\hline
         \multicolumn{3}{l|}{Data distribution}& \multicolumn{5}{c|}{\textbf{IID}}&\multicolumn{5}{c}{\textbf{non-IID}} \\  \hline
        Category&Methods  &Metric  & Benign&F Flip& FOE&S-DFA&Sp-DFA  & Benign&F Flip& FOE&S-DFA&Sp-DFA  \\ \hline
          & & ACC& 65.55&65.01&52.27&55.82&55.51&65.50&64.87&53.30&51.96&54.58\\
          &  FedAvg & AEOD& 0.118&0.193&0.056&0.054&0.103&0.152&0.409&0.021&0.019&0.034 \\ 
          Fairness& & ASPD      &0.113&0.187&0.040&0.073&0.086&0.163&0.393&0.051&0.014&0.051\\ \cline{2-13}
         
         (debias)& &ACC   &65.43&61.23&50.00&50.00&50.00&66.32&61.69&50.00&59.84&53.45\\
         algorithms& FairFed& AEOD 
         &0.081&0.268&\textbf{0.000}&\textbf{0.000}&\textbf{0.000}&0.066&0.330&\textbf{0.000}&0.341&0.076     \\ 
         & & ASPD      &0.098&0.320&\textbf{0.000}&\textbf{0.000}&\textbf{0.000}&0.071&0.308&\textbf{0.000}&0.270&\textbf{0.061}\\ \hline
         
         & &ACC  &65.81&\textbf{65.65}&65.71&65.81&65.65&65.71&\textbf{65.35}&65.37&65.55&65.32\\
         & Median& AEOD
         &0.127&0.165&0.186&0.165&0.211&0.194&0.139&0.262&0.227&0.258    \\ 
          Robust FL& & ASPD      &0.133&0.152&0.189&0.154&0.198&0.186&0.150&0.264&0.233&0.254  \\ \cline{2-13}
         
        algorithms& &ACC  &64.52&63.67&64.18&63.67&63.44&64.62&63.62&64.62&64.63&63.17\\
         for general& FLTrust& AEOD    &0.103&0.196&0.190&0.196&0.197&0.200&0.196&0.240&0.206&0.217   \\ 
         adversarial& & ASPD      &0.102&0.181&0.170&0.181&0.186&0.197&0.200&0.222&0.200&0.217  \\  \cline{2-13}

        attacks& &ACC&63.37&62.54&62.43&60.15&62.25&63.36&62.21&61.47&61.14&62.11\\
        & Class-B FL&AEOD     &0.112&0.186&0.194&0.189&0.172&0.210&0.187&0.192&0.330&0.211     \\ 
        & & ASPD      &0.118&0.198&0.188&0.196&0.184&0.197&0.211&0.215&0.217&0.200  \\ \hline
         
         Robust FL& &ACC&65.35&65.40&51.91&55.61&54.48&65.50&65.29&50.41&50.82&54.27\\
         for fairness& FairGuard&AEOD     &0.116&0.126&0.049&0.037&0.060&0.144&0.116&0.020&0.019&0.040     \\ 
         attacks& & ASPD      &0.115&0.121&0.038&0.065&0.056&0.162&0.135&0.018&0.013&0.061  \\ \hline
         
         \multirow{2}{*}{Robust FL} &FLTrust&ACC   &64.16&63.65&64.18&63.65&63.69&64.23&63.75&64.77&64.75&63.69\\
          \multirow{2}{*}{(hybrid)}& +&AEOD        &0.169&0.178&0.190&0.178&0.194&0.167&0.154&0.280&0.154&0.239  \\ 
         &FairGuard & ASPD      &0.147&0.164&0.170&0.174&0.186&0.148&0.159&0.281&0.159&0.144  \\ \hline
         
         & &ACC & \textbf{65.86}&65.35&\textbf{65.81}&\textbf{65.49}&\textbf{65.81}&\textbf{66.05}&65.17&\textbf{65.62}&\textbf{65.84}&\textbf{65.83}\\
         Ours& GuardFed&AEOD      &\underline{\textbf{0.044}}&\underline{\textbf{0.004}}&\underline{\textbf{0.047}}&\underline{\textbf{0.044}}&\underline{\textbf{0.048}}&\underline{\textbf{0.047}}&\underline{\textbf{0.004}}&\underline{\textbf{0.055}}&\underline{\textbf{0.037}}&\underline{\textbf{0.055}}  \\ 
         & &ASPD      &\underline{\textbf{0.057}}&\underline{\textbf{0.013}}&\underline{\textbf{0.059}}&\underline{\textbf{0.013}}&\underline{\textbf{0.067}}&\underline{\textbf{0.052}}&\underline{\textbf{0.001}}&\underline{\textbf{0.071}}&\underline{\textbf{0.038}}&\underline{\textbf{0.071}}\\ \hline
    \end{tabular} }
    \label{tab:3}
\end{table*}
The experimental results demonstrate that none of the existing fairness-enhanced or robust FL methods can effectively defend against the proposed DFA. Even when combining fairness- and performance-oriented defenses in the Hybrid approach, fairness remains vulnerable. These findings highlight the urgent need for a new robust FL system that can simultaneously ensure both model utility and fairness.

(ii) \textbf{The GuardFed system can effectively defend against both performance- and fairness-harming attacks, including the proposed DFA, and achieves SOTA results in terms of both learning performance and fairness.} When fairness-harming (F Flip) attackers are involved, GuardFed achieves an AEOD of 0.004 on the COMPAS dataset, and 0.004 and 0.011 on the ADULT dataset under IID and non-IID settings, respectively. Similarly, the corresponding ASPD values remain at the lowest levels, with 0.013 and 0.001 on COMPAS, and 0.059 and 0.079 on ADULT. These results indicate GuardFed’s strong capability in preserving fairness under adversarial conditions. GuardFed also demonstrates strong robustness against performance-harming attacks, achieving approximately 65\% testing accuracy on the COMPAS dataset and 82\% on the ADULT dataset when defending against FOE attacks. 

Furthermore, GuardFed consistently maintains state-of-the-art performance in both fairness and utility under the proposed DFA. Regardless of the attack variant (S-DFA or Sp-DFA), it achieves high accuracy (above 65\% on the COMPAS dataset and 82\% on the ADULT dataset) while maintaining low fairness metrics (ASPD and AEOD generally below 0.1). Compared to the hybrid defense, GuardFed achieves approximately 1/10 the fairness index, demonstrating its superior protection against dual-objective attacks. Notably, GuardFed does not compromise between learning performance and fairness; instead, it simultaneously achieves both high utility and strong fairness. This dual advantage makes GuardFed a more practical and deployable framework for real-world smart intelligent compared to existing FL methods. Table  \ref{tab:3} and Table \ref{tab:4} provide comparisons of the learning performance and fairness of the existing robust FL system and our proposed GuardFed in various adversarial environments.

(iii) \textbf{GuardFed achieves competitive performance and fairness in non-adversarial settings.} Specifically, GuardFed achieves testing accuracies of 65.86\% and 66.05\% on the COMPAS dataset, and 83.74\% and 82.12\% on the ADULT dataset across two data distributions, representing the highest or near-highest performance among all compared methods. On the other hand, GuardFed also demonstrates top-tier fairness under benign settings. Except for slightly higher AEOD (0.022) and ASPD (0.096) values compared to FedAvg (AEOD 0.018) and FairFed (ASPD 0.093) in the ADULT IID scenario, GuardFed consistently achieves the lowest or near-zero fairness metrics (AEOD and ASPD) across most benign experimental settings. In contrast, the hybrid method exhibits the most significant bias and records the highest ASPD and AEOD values in most simulated benign scenarios, indicating that simply combining performance- and fairness-enhancing techniques does not necessarily improve fairness under non-adversarial conditions. These performances under the benign environment can be found in Tables \ref{tab:3} and \ref{tab:4}.

\begin{table*}\small
    \centering
        \caption{Comparison of the learning performance of existing robust FL methods and GuardFed on the ADULT dataset. Bold values indicate the best results, while underlines denote fairness scores when the accuracy exceeds the predefined threshold.}
            \makebox[\textwidth][l]{%
    \begin{tabular}{c|c|l|ccccc|ccccc}    
    \hline\hline
         \multicolumn{3}{l|}{Data distribution}& \multicolumn{5}{c|}{\textbf{IID}}&\multicolumn{5}{c}{\textbf{non-IID}} \\  \hline
        Category&Methods  &Metric  & Benign&F Flip& FOE&S-DFA&Sp-DFA  & Benign&F Flip& FOE&S-DFA&Sp-DFA  \\ \hline
         & & ACC &83.05&81.76&76.63&78.30&78.44&82.23&81.51&76.53&78.15&78.34\\
         & FedAvg & AEOD      &\underline{\textbf{0.018}}&0.216&\textbf{0.001}&0.003&\textbf{0.018}&0.055&0.082&0.002&0.007&0.025  \\ 
         Fairness& & ASPD      &0.104&0.121&\textbf{0.011} & 0.029 & 0.037 &0.099&0.145&0.007&0.034&0.040  \\ \cline{2-13}
         
         (debias)& &ACC   &83.19&81.76&75.89&77.00&77.77&82.05&80.63&75.69&77.41&76.01\\
         algorithms& FairFed& AEOD 
         &0.007&0.107&0.004&\textbf{0.002}&0.004&0.056&0.135&0.005&0.012&0.001     \\ 
         & & ASPD      &\underline{\textbf{0.093}}&0.066&0.018 & \textbf{0.022} & \textbf{0.028} &0.101&0.001&\textbf{0.004}&\textbf{0.011}&\textbf{0.005}  \\ \hline
         
         & &ACC  &82.60&82.67&83.18&83.37&83.35&81.50&81.98&81.94&81.96&82.09\\
         & Median& AEOD
         &0.028&0.072&0.018&0.049&0.020&0.065&0.133&0.098&0.065&0.018    \\ 
         Robust FL& & ASPD      &0.095&0.085&0.105 & 0.103 & 0.109  &0.082&\underline{\textbf{0.079}}&\underline{\textbf{0.071}}&\underline{\textbf{0.078}}&0.088  \\ \cline{2-13}
         
         algorithms& &ACC  &82.39&82.56&82.64&82.73&81.80&81.98&82.14&82.32&82.23&81.74\\
         for general& FLTrust& AEOD               &0.274&0.309&0.298&0.327&0.214&0.248&0.251&0.258&0.267&0.220    \\ 
         adversarial& & ASPD      &0.196&0.281&0.189 & 0.252 & 0.146  &0.161&0.152&0.166&0.172&0.161  \\ \cline{2-13}

         attacks& &ACC&81.92&82.13&81.41&80.98&80.02&82.35&80.94&81.68&79.45&78.72\\
         & Class-B FL&AEOD     &0.298&0.294&0.282&0.318&0.239&0.254&0.260&0.256&0.271&0.239     \\ 
          & & ASPD      &0.198&0.279&0.194 & 0.260 & 0.133  &0.162&0.160&0.169&0.187&0.168  \\ \hline

         Robust FL& &ACC&82.93&82.45&76.34&78.97&79.32&\textbf{83.33}&\textbf{82.64}&76.53&79.16&79.21\\
         for fairness& FairGuard&AEOD     &0.003&0.091&0.002&0.037&0.023&0.142&0.026&\textbf{0.001}&0.006&0.023     \\ 
         attacks& & ASPD      &0.107&0.073&0.009 & 0.026 & 0.035  &0.141&0.081&0.009&0.033&0.038  \\ \hline
         
         \multirow{2}{*}{Robust FL}&FLTrust &ACC   &82.69&81.35&82.80&82.44&82.00&82.02&82.04&\textbf{82.35}&82.46&81.92\\
         \multirow{2}{*}{(hybrid)}& + &AEOD        &0.313&0.442&0.341&0.377&0.265&0.267&0.244&0.276&0.275&0.250  \\ 
         &FairGuard & ASPD      &0.196&0.281&0.189 & 0.252 & 0.146  &0.160&0.152&0.166&0.172&0.161  \\ \hline
         
         & &ACC &\textbf{83.74}&\textbf{83.83}&\textbf{83.77}&\textbf{83.73}&\textbf{83.72}&82.12&81.99&81.58&\textbf{82.53}&\textbf{82.58}\\
         Ours& GuardFed&AEOD      &0.022&\underline{\textbf{0.004}}&\underline{\textbf{0.053}}&\underline{\textbf{0.026}}&\underline{\textbf{0.051}}&\underline{\textbf{0.033}}&\underline{\textbf{0.011}}&\underline{\textbf{0.015}}&\underline{\textbf{0.006}}&\underline{\textbf{0.015}}    \\ 
         & &ASPD      &0.096&\underline{\textbf{0.059}}&\underline{\textbf{0.090}}&\underline{\textbf{0.071}}&\underline{\textbf{0.090}}&\underline{\textbf{0.081}}&\underline{\textbf{0.079}}&0.084&0.093&\underline{ \textbf{0.084}}   \\ \hline
    \end{tabular} }

    \label{tab:4}
\end{table*}

\subsection{Exploratory Study and Discussion}
To provide a more comprehensive understanding of GuardFed’s stability, we investigate the effects of key hyperparameters, including the balancing coefficient $\tau$ (from Formula~\ref{16}) and the size of the server-side dataset. We also extend the experiments to evaluate GuardFed's performance under different attacker ratios and configuration settings. In the end of this Section, we summarize the limitations of this work and discuss directions for future research. Given that the non-IID setting with Sp-DFA represents the most challenging scenario, we primarily monitor Accuracy and AEOD under this configuration using the ADULT dataset. The COMPAS dataset is additionally included when evaluating the impact of different attacker ratios.

\textbf{(i) Impact of $\tau$ on Trust Score computation.}
The parameter $\tau$ acts as a balancing coefficient between the fairness index and the deviation index in the trust score formulation. It determines the relative importance placed on fairness violations versus utility degradation, thereby influencing how the system evaluates and filters potentially harmful client updates. As $\tau$ decreases, the trust score increasingly relies on the deviation index alone, with fairness considerations diminishing; when $\tau = 0$, the trust score depends solely on deviation.

\begin{figure*}
    \centering
    \includegraphics[width=1.0\linewidth]{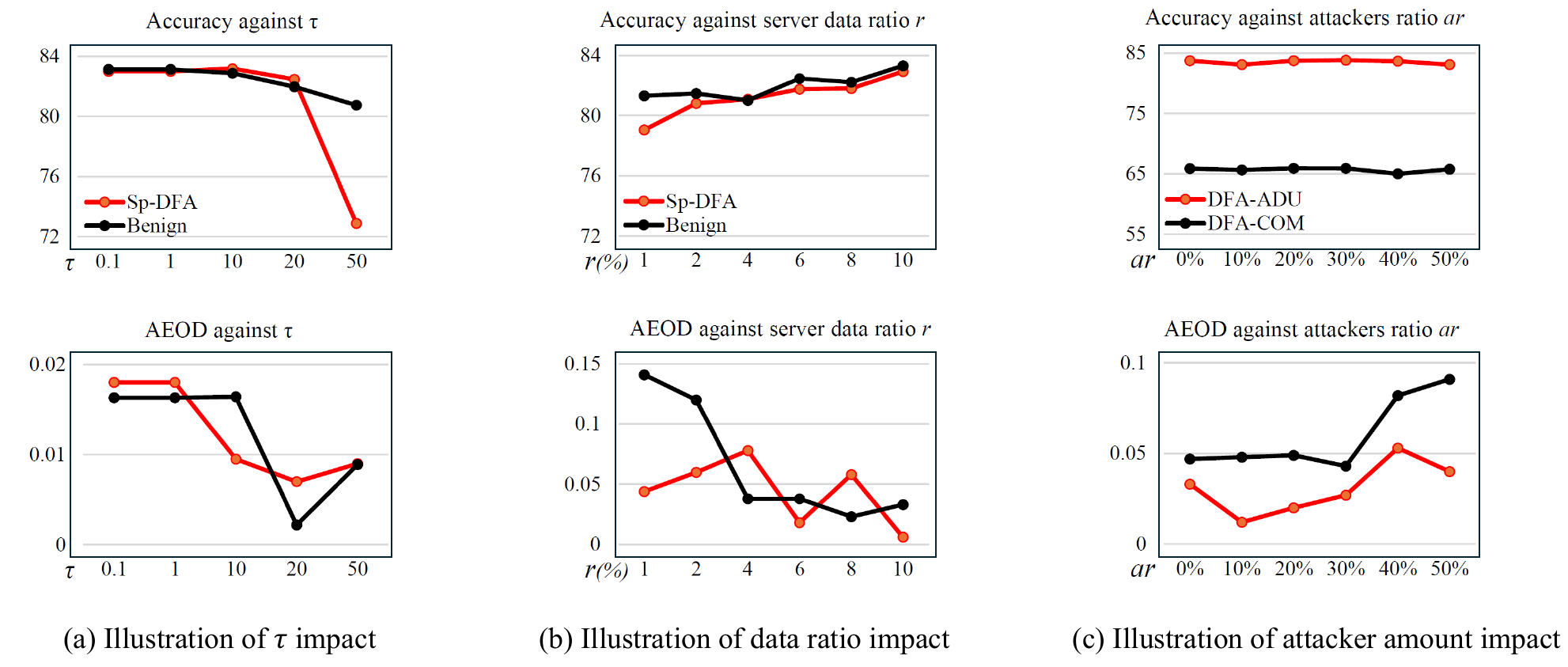}
\caption{Illustration of model accuracy and AEOD under different hyperparameters and adversarial settings }
    \label{c}
\end{figure*}

Experimental results (Figure \ref{c}, subplots-a) confirm that increasing $\tau$ from 0.1 to 50 leads to improved fairness (reflected by decreasing AEOD) but results in reduced learning performance. In extreme cases, such as when Sp-DFA attackers are present and $\tau=50$: the model completely breaks down, highlighting a clear trade-off between fairness and utility. Therefore, $\tau$ should be tuned according to task-specific priorities: if fairness is critical, a relatively large $\tau$ is appropriate; otherwise, a smaller value may be preferred to preserve model performance. However, it is important to note that although $\tau$ introduces a trade-off between fairness and learning performance, this does not imply that GuardFed sacrifices both aspects, as observed in many existing FL methods. As illustrated in Figure~\ref{c} and Table~\ref{tab:4}, the degradation in either dimension remains marginal. GuardFed consistently achieves the best overall performance across a wide range of $\tau$ settings, outperforming other FL methods.

\textbf{(ii) Impact of the server dataset size.}
To provide a better understanding of the impact of server dataset size, we conduct experiments using varying proportions of server data, ranging from 1\% (manually collected root data only) to 10\% of the overall dataset size. As shown in the subplots-b of Figure~\ref{c}, both the testing accuracy and AEOD improve as the server dataset size increases, indicating enhanced learning performance and fairness under both benign and adversarial conditions.

However, compared to accuracy, AEOD is less sensitive to the increase in server data size. While the overall trend of AEOD is downward, it reaches a relatively low level when the server data size exceeds 4\%, and further increases only result in marginal gains. For instance, under the presence of Sp-DFA attackers, increasing the server data size from 4\% to 10\% reduces AEOD from approximately 0.07 to 0.01. In benign settings, AEOD decreases from around 0.04 to 0.025. Although these reductions demonstrate continued improvement, GuardFed already achieves significantly lower AEOD compared to existing FL methods, even with a smaller server dataset.

\textbf{(iii) Impact of the adversary ratio.}
In real-world federated learning scenarios, the proportion of adversarial participants can vary significantly. To evaluate the generalizability of GuardFed under varying threat levels, we conduct experiments with attacker ratios ranging from 0\% to 50\%. Here, 50\% attacker ratio represents an extreme and challenging condition, while 20\% is widely regarded as a realistic upper bound. As shown in Figure~\ref{c} (subplots-c), GuardFed consistently maintains strong performance in both accuracy and fairness. Even when half of the participants are Sp-DFA attackers, GuardFed achieves stable results on both the COMPAS and ADULT datasets. Specifically, the testing accuracy fluctuates by less than 1.5\% as the attacker ratio increases. Although AEOD slightly rises when the adversary proportion exceeds 30\%, peaking at approximately 0.08 on ADULT and 0.05 on COMPAS, these values remain lower than the fairness scores of existing FL methods even in benign settings. For instance, FairFed yields AEOD values of 0.066 on COMPAS and 0.056 on ADULT without any attacks. These results confirm GuardFed’s effectiveness in preserving both learning performance and fairness, even under high adversarial pressure.

\textbf{(iv) Limitations and Future Works.} While DFA and GuardFed have demonstrated effectiveness in attacking and defending federated learning systems, several limitations remain that warrant further investigation.
First, most existing studies on fairness-aware federated learning, including this work, focus primarily on tabular learning tasks. When extending to image-based or multimodal recognition tasks, the effectiveness of current attack and defense strategies remains largely unknown. Therefore, adapting the proposed framework to vision and multimodal federated learning represents an interesting and meaningful direction for future research.

Second, the present study mainly considers binary sensitive attributes, such as gender (male/female). However, in real-world applications, sensitive variables are often multi-valued (e.g., age, income level) where fairness definitions and evaluation mechanisms for ternary or higher-order attributes have yet to be systematically explored. Investigating fairness-aware federated learning under such complex sensitive characteristics will thus be an important avenue for future work.

\section{Conclusion} \label{S7}
In this work, we introduce the Dual-Facet Attack (DFA), a novel threat model that simultaneously compromises both utility and fairness in federated learning. To counter this challenge, we propose GuardFed, a self-adaptive defense framework that leverages a small amount of clean server data to construct a fairness-aware reference model. GuardFed jointly evaluates client updates from utility and fairness perspectives and selectively aggregates those with high trust scores.

Extensive experiments on real-world datasets across varying distributions and attacker ratios show that GuardFed robustly maintains both high accuracy and fairness, even under severe adversarial conditions. These findings demonstrate the strong robustness and practical value of GuardFed, highlighting its potential as a generalizable and trustworthy framework for secure and fair federated learning.

\bibliographystyle{IEEEtran}
\bibliography{paper}
\end{document}